\documentclass[11pt,a4paper]{article}
\usepackage{times,latexsym}
\usepackage{url}
\usepackage[T1]{fontenc}
\usepackage{varwidth}
\usepackage{multirow}
\usepackage{bm}
\usepackage{amsthm}
\usepackage{adjustbox}
\usepackage{amsmath}
\usepackage{amsmath}
\DeclareMathOperator*{\argmin}{argmin} 
\usepackage{amssymb}
\usepackage{makecell}
\usepackage[linesnumbered,ruled,vlined]{algorithm2e}
\usepackage{float}
\usepackage{xcolor}
\usepackage{graphicx}
\graphicspath{ {./Images/} }
\usepackage{caption}
\usepackage{subcaption}
\usepackage{enumitem}
\restylefloat{table}
\usepackage{xfrac}
\definecolor{blue}{rgb}{0,0,0}
\usepackage[acceptedWithA, nohyperref]{tacl2018v2}

\usepackage{xspace,mfirstuc,tabulary}

\newif\iftaclinstructions
\taclinstructionsfalse % AUTHORS: do NOT set this to true
\iftaclinstructions
\renewcommand{\anonsubtext}{(No author info supplied here, for consistency with
TACL-submission anonymization requirements)}
\newcommand{\instr}
\fi

\iftaclpubformat % this "if" is set by the choice of options

\else

\fi

\usepackage{amsthm}

\title{Nurse is Closer to Woman than Surgeon? Mitigating Gender-Biased Proximities in Word Embeddings}

\newcommand*{\affaddr}[1]{#1} % No op here. Customize it for different styles.
\newcommand*{\affmark}[1][*]{\textsuperscript{#1}}
\newcommand*{\email}[1]{\texttt{#1}}

\author{
Vaibhav Kumar \affmark[1]\thanks{\hspace{1mm}Authors have contributed equally.} \hspace{0.3cm} Tenzin Singhay Bhotia \affmark[1]\footnotemark[1]\hspace{0.3cm} Vaibhav Kumar \affmark[1]\footnotemark[1]\hspace{0.3cm} Tanmoy Chakraborty\affmark[2]\\
\affaddr{\affmark[1]Delhi Technological University, New Delhi, India}\\
\affaddr{\affmark[2]IIIT-Delhi, India}\\
\email{\affmark[1]\{kumar.vaibhav1o1, tenzinbhotia0, vaibhavk992\}@gmail.com}\\
\email{\affmark[2]tanmoy@iiitd.ac.in}\\
}

\begin{document}

\maketitle

\begin{abstract}

Word embeddings are the standard model for semantic and syntactic representations of words. Unfortunately, these models have been shown to exhibit undesirable word associations resulting from gender, racial, and religious biases. Existing post-processing methods for debiasing word embeddings are unable to mitigate gender bias hidden in the spatial arrangement of word vectors. In this paper, we propose {\bf RAN-Debias}, a novel gender debiasing methodology which not only eliminates the bias present in a word vector but also alters the spatial distribution of its neighbouring vectors, achieving a bias-free setting while maintaining minimal semantic offset. We also propose a new bias evaluation metric - {\bf Gender-based Illicit Proximity Estimate} (GIPE), which measures the extent of undue proximity in word vectors resulting from the presence of gender-based predilections. Experiments based on a suite of evaluation metrics show that RAN-Debias significantly outperforms the state-of-the-art in reducing proximity bias (GIPE) by at least 42.02\%. It also reduces direct bias, adding minimal semantic disturbance, and achieves the best performance in a downstream application task (coreference resolution).

\end{abstract}

\section {Introduction}

\noindent Word embedding methods \cite{devlin2018bert,mikolov2013efficient,pennington2014GloVe} have been staggeringly successful in mapping the semantic space of words to a space of real-valued vectors, capturing both semantic and syntactic relationships. 
However, as recent research has shown, word embeddings also possess a spectrum of biases related to gender \cite{bolukbasi2016man,hoyle-etal-2019}, race, and religion \cite{manzini2019black,10.1145}. \citet{bolukbasi2016man} showed that there is a disparity in the association of professions with gender. For instance, while women are associated more closely with `receptionist' and `nurse', men are associated more closely with `doctor' and `engineer'. Similarly, a word embedding model trained on data from a popular social media platform generates analogies such as ``Muslim is to terrorist as Christian is to civilian'' \cite{manzini2019black}. Therefore, given the large scale use of word embeddings, it becomes cardinal to remove the manifestation of biases. In this work, we focus on mitigating gender bias from pre-trained word embeddings.

\begin{table}[t!]
\centering
\scalebox{0.8}{
\begin{tabular}{|l| l|} 
 \hline
 \multicolumn{1}{|c|}{\textbf{Word}} & \multicolumn{1}{c|}{\textbf{Neighbours}} \\  
 \hline
 nurse & mother\textsubscript{12}, woman\textsubscript{24}, filipina\textsubscript{31} \\ 
 receptionist & housekeeper\textsubscript{9}, hairdresser\textsubscript{15}, prostitute\textsubscript{69} \\
 prostitute & housekeeper\textsubscript{19}, hairdresser\textsubscript{41}, babysitter\textsubscript{44} \\
schoolteacher & homemaker\textsubscript{2}, housewife\textsubscript{4}, waitress\textsubscript{8}\\
 \hline
\end{tabular}}
\caption{Words and their neighbours extracted using GloVe \cite{pennington2014GloVe}. Subscript indicates the rank of the neighbour.}
\label{table:1}
\vspace{-0mm}
\end{table}

As shown in Table \ref{table:1}, \textcolor{blue}{the high degree of similarity between gender-biased words largely results from their individual proclivity towards a particular notion (gender in this case) rather than from empirical utility; we refer to such proximities as ``illicit proximities''.} Existing debiasing methods \cite{bolukbasi2016man,kaneko2019gender} are primarily concerned with debiasing a word vector by minimising its projection on the gender direction. Although they successfully mitigate direct bias for a word, they tend to ignore the relationship between a gender-neutral word vector and its neighbours, thus failing to remove the gender bias encoded as illicit proximities between words \cite{gonen2019lipstick,williams-etal-2019-quantifying}. For the sake of brevity,  we refer to `gender-based illicit proximities' as `illicit  proximities' in  the  rest of the paper.

To account for these problems, we propose a post-processing based debiasing scheme for non-contextual word embeddings, called {\bf RAN-Debias} ({\bf R}epulsion, {\bf A}ttraction, and {\bf N}eutralization based {\bf Debias}ing). RAN-Debias not only minimizes the projection of gender-biased word vectors on the gender direction but also reduces the semantic similarity with neighbouring word vectors having illicit proximities. We also propose \textbf{KBC} (\textbf{K}nowledge \textbf{B}ased \textbf{C}lassifier), a word classification algorithm for selecting the set of words to be debiased. KBC utilizes a set of existing lexical knowledge bases to maximize classification accuracy. Additionally, we propose a metric - {\bf Gender-based Illicit Proximity Estimate (GIPE)}, which quantifies gender bias in the embedding space resulting from the presence of illicit proximities between word vectors.

We evaluate debiasing efficacy on various evaluation metrics. For the gender relational analogy test on the SemBias dataset \cite{zhao2018learning}, RAN-GloVe (RAN-Debias applied to GloVe word embedding) outperforms the next best baseline GN-GloVe (debiasing method proposed by \citet{zhao2018learning}) by 21.4\% in gender-stereotype type. RAN-Debias also outperforms the best baseline by at least 42.02\% in terms of GIPE.
Furthermore, the performance of RAN-GloVe on word similarity and analogy tasks on a number of benchmark datasets indicates the addition of minimal semantic disturbance. In short, our major contributions\footnote{The code and data are released at \url{https://github.com/TimeTraveller-San/RAN-Debias}} can be summarized as follows:
\\
\noindent$\bullet$ We provide a knowledge based method (KBC) for classifying words to be debiased.
\\
\noindent$\bullet$ We introduce RAN-Debias, a novel approach to reduce both direct and gender-based proximity biases in word embeddings.
\\
\noindent$\bullet$ We propose GIPE, a novel metric to measure the extent of undue proximities in word embeddings.

\section{Related Work}

\subsection{Gender Bias in Word Embedding Models}
\citet{caliskan2017semantics} highlighted that human-like semantic biases are reflected through word embeddings (such as GloVe \cite{pennington2014GloVe}) of ordinary language. They also introduced the Word Embedding Association Test (WEAT) for measuring bias in word embeddings. The authors showed a strong presence of biases in pre-trained word vectors. In addition to gender, they also identified bias related to race.
For instance, European-American names are more associated with pleasant terms as compared to African-American names.

In the following subsections, we discuss existing gender debiasing methods based on their mode of operation. Methods that operate on pre-trained word embeddings are known as {\em post-processing methods}, while those which aim to retrain word embeddings by either introducing corpus-level changes or modifying the training objective are known as {\em learning-based methods}.

\subsection{Debiasing Methods (Post-processing)}
\citet{bolukbasi2016man} extensively studied gender bias in word embeddings and proposed two debiasing strategies -- `hard debias' and `soft debias'. Hard debias algorithm first determines the direction which captures the gender information in the word embedding space using the difference vectors (e.g., $\vec{he}-\vec{she}$). It then transforms each word vector $\vec{w}$ to be debiased such that it becomes perpendicular to the gender direction (neutralization). Further, for a given set of word pairs (equalization set), it modifies each pair such that $\vec{w}$ becomes equidistant to each word in the pair (equalization). On the other hand, the soft debias algorithm applies a linear transformation to word vectors, which preserves pairwise inner products amongst all the word vectors while limiting the projection of gender-neutral words on the gender direction. The authors showed that the former performs better for debiasing than the latter. However, to determine the set of words for debiasing, a support vector machine (SVM) classifier is used, which is trained on a small set of seed words. This makes the accuracy of the approach highly dependent on the generalization of the classifier to all remaining words in the vocabulary.

 \citet{kaneko2019gender} proposed a post-processing step in which the given vocabulary is split into four classes -- \textcolor{blue}{non-discriminative female-biased words (e.g., `bikini', `lipstick'), non-discriminative male-biased words (e.g., `beard', `moustache'), gender-neutral words (e.g., `meal', `memory'), and stereotypical words (e.g., `stance librarian', `doctor').} A set of seed words is then used for each of the categories to train an embedding using an encoder in a denoising autoencoder, such that gender-related biases from stereotypical words are removed, while \textcolor{blue}{preserving feminine information for non-discriminative female-biased words, masculine information for non-discriminative male-biased words, and neutrality of the gender-neutral words.} The use of the correct set of seed words is critical for the approach. Moreover, inappropriate associations between words (such as `nurse' and `receptionist') may persist.

 \citet{gonen2019lipstick} showed that current approaches \cite{bolukbasi2016man,zhao2018learning}, which depend on gender direction for the definition of gender bias and directly target it for the mitigation process, end up hiding the bias rather than reduce it. The relative spatial distribution of word vectors before and after debiasing is similar, and bias-related information can still be recovered. 

\citet{ethayarajh2019understanding} provided theoretical proof for hard debias \cite{bolukbasi2016man} and discussed the theoretical flaws in WEAT by showing that it systematically overestimates gender bias in word embeddings. The authors presented an alternate gender bias measure, called RIPA (Relational Inner Product Association), that quantifies gender bias using gender direction. Further, they illustrated that vocabulary selection for gender debiasing is as crucial as the debiasing procedure.

\citet{zhou-etal-2019-examining} investigated the presence of gender bias in bilingual word embeddings and languages which have grammatical gender (such as Spanish and French). Further, they defined semantic gender direction and grammatical gender direction used for quantifying and mitigating gender bias. In this paper, we only focus on languages that have non-gendered grammar (e.g., English). Our method can be applied to any such language. 

\subsection{Debiasing Methods (Learning-based)}

\citet{zhao2018learning} developed a word vector training approach, called Gender-Neutral Global Vectors (GN-GloVe) based on the modification of GloVe. They proposed a modified objective function that aims to confine gender-related information to a sub-vector. During the optimization process, the objective function of GloVe is minimized while simultaneously, the square of Euclidean distance between the gender-related sub-vectors is maximized. Further, it is emphasized that the representation of gender-neutral words is perpendicular to the gender direction. Being a retraining approach, this method cannot be used on pre-trained word embeddings.

\citet{lu2018gender} proposed a counterfactual data augmentation (CDA) approach to show that gender bias in language modeling and coreference resolution can be mitigated through balancing the corpus by exchanging gender pairs like `she' and `he' or `mother' and `father'. Similarly, \citet{maudslay2019s} proposed a learning-based approach with two enhancements to CDA --  a counterfactual data substitution method which makes substitutions with a probability of $0.5$ and a method for processing first names based upon bipartite graph matching.

\citet{bordia2019identifying} proposed a gender-bias reduction method for word-level language models. They introduced a regularization term that penalizes the projection of word embeddings on the gender direction. Further, they proposed metrics to measure bias at embedding and corpus level. Their study revealed considerable gender bias in Penn
Treebank \cite{marcus1993building} and WikiText-2 \cite{merity2017regularizing}.

\subsection{Word Embeddings Specialization}
\citet{mrkvsic2017semantic} defined semantic specialization as the process of refining word  vectors to improve the semantic content. Similar to the debiasing procedures, semantic specialization procedures can also be divided into {\em post-processing} \cite{ono2015word,faruqui2014improving} and {\em learning-based} \cite{rothe2015autoextend,mrkvsic2016counter,nguyen2016integrating} approaches. The performance of post-processing based approaches is shown to be better than learning-based approaches \cite{mrkvsic2017semantic}.

Similar to the `repulsion' and `attraction' terminologies used in RAN-Debias, \citet{mrkvsic2017semantic} defined ATTRACT-REPEL algorithm, a post-processing semantic specialization process which uses antonymy and synonymy constraints drawn from lexical resources. 
Although it is superficially similar to RAN-Debias, there are a number of differences between the two approaches. \textcolor{blue}{Firstly, the \textsc{Attract}-\textsc{Repel} algorithm operates over mini-batches of synonym and antonym pairs, while RAN-Debias operates on a set containing gender-neutral and gender-biased words. Secondly, the `attract' and `repel' terms carry different meanings with respect to the algorithms. In \textsc{Attract}-\textsc{Repel}, for each of the pairs in the mini-batches of synonyms and antonyms, negative examples are chosen. The algorithm then forces synonymous pairs to be closer to each other (attract) than from their negative examples and antonymous pairs further away from each other (repel) than from their negative examples. On the other hand, for a given word vector, RAN-Debias forces it away from its neighbouring word vectors (repel) which have a high indirect bias while simultaneously forcing the post-processed word vector and the original word vector together (attract) to preserve its semantic properties.}

\begin{table}[t]
    \centering
    \scalebox{0.9}{
    \begin{tabular}{|c|l|}
         \hline
         {\bf Notation} & {\bf Denotation}  \\\hline
         $\vec{w}$ & Vector corresponding to a word $w$
         \\\hline
         $\vec{w_d}$ & Debiased version of $\vec{w}$
         \\\hline
         $V$ & Vocabulary set
         \\\hline
         $V_p$ & The set of words which are preserved\\& during the debiasing procedure
         \\\hline
         $V_d$ & The set of words which are subjected\\& to the debiasing procedure
         \\\hline
         $D$ & Set of dictionaries
         \\\hline
         $d_{i}$ & A particular dictionary from the set $D$
         \\\hline
         $\vec{g}$ & Gender direction
         \\\hline
         $D_b(\vec{w})$ & Direct bias of a word $w$. 
         \\\hline
         $\beta(\vec{w_{1}},\vec{w_{2}})$ & Indirect bias between a pair of words \\&$w_{1}$ and $w_{2}$.
         \\\hline
         $\eta(\vec{w})$ & Gender-based proximity bias
         \\ & of a word $w$
         \\\hline
         $N_{w}$ & Set of neighbouring words of a word $w$ 
         \\\hline         
         $F_r(\vec{w_d})$ & Repulsion objective function
         \\\hline
         $F_a(\vec{w_d})$ & Attraction objective function
         \\\hline
         $F_n(\vec{w_d})$ & Neutralization objective function
         \\\hline
         $F(\vec{w_d})$ & Multi-objective optimization function
         \\\hline
         $KBC$ & Knowledge Based Classifier
         \\\hline         
         $BBN$ & Bias Based Network
         \\\hline
         $GIPE$ & Gender-based Illicit Proximity
         \\ & Estimate
         \\\hline
    \end{tabular}
    }
    \caption{Important notations and denotations.}
    \label{tab:notation}
\end{table}

\section{Proposed Approach}

Given a set of pre-trained word vectors $\left \{\vec{w}_{i} \right \}_{i=1}^{|V| }$ over a vocabulary set \textit{V},
we aim to create a transformation $\left \{\vec{w}_{i} \right \}_{i=1}^{|V| } \rightarrow  \{\vec{w^{'}}_{i}  \}_{i=1}^{|V| }$ such that the stereotypical gender information present in the resulting embedding set are minimized with minimal semantic offset. We first define the categories into which each word \( w \in V\) is classified in a mutually exclusive manner. Table \ref{tab:notation} summarizes important notations used throughout the paper.

\noindent$\bullet$ \textbf{Preserve set $\left(V_{p} \right)$}: This set consists of words for which gender carries semantic importance; such as names, gendered pronouns and words like `beard' and `bikini' which have a meaning closely associated with gender. In addition, words which are non-alphabetic are also included as debiasing them will be of no practical utility. 
\\
\noindent$\bullet$ \textbf{Debias set ($V_{d})$}:
This set consists of all the words in the vocabulary which are not present in $V_{p}$. These words are expected to be gender-neutral in nature and hence subjected to debiasing procedure. \textcolor{blue}{Note that $V_{d}$ not only consists of gender-stereotypical words (e.g., `nurse', `warrior', `receptionist', etc.), but also gender-neutral words (e.g., `sky', `table', `keyboard', etc.).}

\subsection{Word Classification Methodology}
Prior to the explanation of our method, we present the limitations of previous approaches for word classification. \citet{bolukbasi2016man} trained a linear SVM using a set of gender-specific seed words, which is then generalized on the whole embedding set to identify other gender-specific words. However, such methods rely on training a supervised classifier on word vectors, which are themselves gender-biased. Because such classifiers are trained on biased data, they catch onto the underlying gender-bias cues and often misclassify words. For instance, the SVM classifier trained by ~\citet{bolukbasi2016man} misclassifies the word `blondes' as gender-specific, among others. Further, we empirically show the inability of a supervised classifier (SVM) to generalize over the whole embedding using various metrics in Table~\ref{table:kfbc}.

Taking into consideration this limitation, we propose Knowledge Based Classifier (KBC) that relies on knowledge bases instead of word embeddings, thereby circumventing the addition of bias in the classification procedure. Moreover, unlike RIPA \cite{ethayarajh2019understanding}, our approach does not rely on creating a biased direction that may be difficult to determine. Essentially, KBC relies on the following assumption.

\theoremstyle{plain}
\newtheorem{assumption}{Assumption}
\let\origtheassumption\theassumption

\begin{assumption}\label{as:assump1}
If there exists a dictionary $d$ such that it stores a definition $d[w]$ corresponding to a word $w$, then  $w$ can be defined as gender-specific or not based on the existence or absence of a gender-specific reference $s \in seed$ in the definition $d[w]$, where the set $seed$  consists of gender-specific references such as \{`man', `woman', `boy', `girl'\}.
\end{assumption}

\begin{algorithm}[!t]\small
\DontPrintSemicolon
\SetKwInOut{Input}{Input}
\SetKwInOut{Output}{Output}
\Input{$V$: vocabulary set,
       $isnonaphabetic(w)$: checks for non-alphabetic words \newline
       $seed$: set of gender-specific words\newline
       $stw$: set of stop words \newline
       $names$: set of gender-specific names \newline
       $D$: set of dictionaries, where for a dictionary $d_{i} \in D$,
       \(d_{i}[w]\) represents the definition of a word \(w\).
       }
\Output{  $V_{p}$: set of words that will be preserved,\newline 
          $V_{d}$: set of words that will be debiased
    }

$V_{p} = \{ \}, V_{d} = \{ \}$\\
 \For{$w \in V$}{
    \uIf{$w \in stw$ or isnonalphabetic(w)}{
         $V_{p} \leftarrow V_{p} \cup \{ w \}$}
    \uElseIf{\(w \in names \cup seed \)}{
         $V_{p} \leftarrow V_{p} \cup \{ w \}$}
    \uElseIf {\(|\{ d_{i} : d_{i} \in D \mathbin{\&} \newline w \in d_{i} \mathbin{\&} \exists s : s \in seed \cap d_{i}[w]\}| > |D|/2 \)}{ 
    $V_{p} \leftarrow V_{p} \cup \{ w \}$}
}
$V_{d} \leftarrow  V_{d} \cup\{ w : w \in  V \setminus V_{p} \} $\;

\Return{$V_{p}, V_{d}$}\;
\caption{Knowledge Based Classifier (KBC)}
\label{algo:kbc}
\end{algorithm}

Algorithm \ref{algo:kbc} formally explains KBC. We denote each {\em if condition} as a stage and explain it below:

\noindent$\bullet$ \textbf{Stage 1: } This stage classifies all stop words and non-alphabetic words as \(V_{p}\). Debiasing such words serve no practical utility; hence we preserve them.
    
\noindent$\bullet$  \textbf{Stage 2: } This stage classifies all words that  belong to either $names$ set or $seed$ set as \(V_{p}\). Set $names$ is collected from open source knowledge base\footnote{\url{https://github.com/ganoninc/fb-gender-json}}. Set $seed$ consists of gender-specific reference terms. We preserve names, as they hold important gender information \cite{pilcher2017names}.

\noindent$\bullet$  \textbf{Stage 3: } This stage uses a collection of dictionaries to determine if a word is gender-specific using Assumption \ref{as:assump1}. To counter the effect of biased definition arising from any particular dictionary, we make a decision based upon the consensus of all dictionaries. A word is classified as gender-specific and added to $V_{p}$ if and only if more than half of the dictionaries classify it as gender-specific. In our experiments, we employ WordNet \cite{miller1995wordnet} and the Oxford dictionary. As pointed out by \citet{bolukbasi2016man}, WordNet consists of few definitions which are gender-biased such as the definition of `vest'; therefore, by utilizing our approach, we counter such cases as the final decision is based upon consensus.

The remaining words that are not preserved by KBC are categorized into $V_{d}$. It is the set of words which are debiased by RAN-Debias later.

\subsection{Types of Gender Bias}\label{typesGB}
First, we briefly explain two types of gender bias as defined by \citet{bolukbasi2016man} and then introduce a new type of gender bias resulting from illicit proximities in word embedding space.
     
\noindent$\bullet$  \textbf{Direct Bias} (\(D_{b}\)): For a word $w$, the direct bias is defined by,
\begin{equation*}
\label{eqn:direct}
D_{b}(\vec{w},\vec{g}) = |cos(\vec{w}, \vec{g})|^{c}
% \frac{\vec{w}.\vec{g}}{|\vec{w}|.|\vec{g}|}
\end{equation*}
where, $\vec{g}$ is the gender direction measured by taking the first principal component from the principal component analysis of ten gender pair difference vectors, such as \((\vec{he} - \vec{she})\) as mentioned in \cite{bolukbasi2016man}, and \(c\) represents the strictness of measuring bias. 

\noindent$\bullet$ \textbf{Indirect Bias (\(\beta\))}: 
The indirect bias between a given  pair of words \(w\) and \(v\) is defined by,
\begin{equation*}
\label{eqn:Indirect}
\beta (\vec{w}, \vec{v}) = \frac{\left ( \vec{w}.\vec{v} - cos(\vec{w}_{\perp}, \vec{v}_{\perp}) \right )}{\vec{w}.\vec{v}}
\end{equation*}
Here, $\vec{w}$ and $\vec{v}$ are normalized. \(\vec{w}_{\perp}\) is orthogonal to the gender direction $\vec{g}$: \(\vec{w}_{\perp} = \vec{w} - \vec{w}_{g}\), and \(\vec{w}_{g}\) is the contribution from gender: \(\vec{w}_{g} =(\vec{w}.\vec{g})\vec{g}\). Indirect bias measures the change in the inner product of two word vectors as a proportion of the earlier inner product after projecting out the gender direction from both the vectors. A higher indirect bias between two words indicates a strong association due to gender.
 
\noindent$\bullet$   \textbf{Gender-based Proximity Bias ($\eta$)}:
\citet{gonen2019lipstick} observed that the existing debiasing methods are unable to completely debias word embeddings because the relative spatial distribution of word embeddings after the debiasing process still encapsulates bias-related information. Therefore, we propose gender-based proximity bias that aims to capture the illicit proximities arising between a word and its closest \textit{k} neighbours due to gender-based constructs.
For a given word \(w_{i}\ \in  V_{d}\),  
the gender-based proximity bias $\eta_{w_{i}}$ is defined as:
\begin{equation}\label{eqn:gbpb}
\eta _{w_{i}} = \frac{|N^b_{w{_i}}|}{|N_{w_i}|}
\end{equation}
where,  

\noindent$N_{w{_i}} =  \underset{V' : |V'|=k}{\text{argmax}} \left (\cos(\vec{w_{i}}, \vec{w_{k}}): w_{k} \in V' \subseteq V\right )$, 
    $N^b_{w{_i}}=\left \{w_{i}:\beta(\vec{w_i}, \vec{w_k}) > \theta_s, \ w_{k} \in N_{w_{i}} \right \}
    $, and \(\theta_s\) is a threshold for indirect bias. 
    
The intuition behind this is as follows. The set \(N_{w{_i}}\) consists of the top $k$ neighbours of \(w_i\) calculated by finding the word vectors having the maximum cosine similarity with \(w_i\). Further, \(N^b_{w{_i}}\subseteq N_{w_i}\) is the set of neighbours having indirect bias \(\beta\) greater than a threshold \(\theta_s\), which is a hyperparameter that controls neighbour deselection on the basis of indirect bias. \textcolor{blue}{The lower is the value of $\theta_s$, the higher is the cardinality of set \(N^b_{w{_i}}\).} A high value of \(|N^b_{w{_i}}|\)  compared to  \(|N_{w_i}|\) indicates that the neighbourhood of the word is gender-biased.
 
\subsection{Proposed Method -- RAN-Debias}

We propose a multi-objective optimization based solution to mitigate both direct\footnote{{\color{blue}Though not done explicitly, reducing direct bias also reduces indirect bias as stated by \citet{bolukbasi2016man}}.} and gender-based proximity bias while adding minimal impact to the semantic and analogical properties of the word embedding. For each word \(w \in V_d\) and its vector \( \vec{w} \in \mathbb{R}^h \), where \(h\) is the embedding dimension, we find its debiased counterpart \( \vec{w_d} \in \mathbb{R}^h \) by solving the following multi-objective optimization problem:
\begin{equation}\label{eqn:obj}
    \argmin_{\vec{w_d}}\big(F_r(\vec{w_d}), F_a(\vec{w_d}), F_n(\vec{w_d})\big)
\end{equation}
We solve this by formulating a single objective \(F(\vec{w_d})\) and scalarizing the set of objectives using the weighted sum method as follows:
\begin{equation}
\begin{split}
    &F(\vec{w_d}) \ = \ \lambda_{1}.F_r(\vec{w_d}) +\lambda_{2}.F_a(\vec{w_d}) + \lambda_{3}.F_n(\vec{w_d})\\
&\text{such that\ }   \lambda_i \in [0, 1] \: \text{and} \:  \sum _i \lambda_i=1    
\end{split}
\label{eq:objective}
\end{equation}
\(F(\vec{w_d})\) is minimized using the adam \cite{kingma2014adam} optimized gradient descent to obtain the optimal debiased embedding \(\vec{w_d}\). 

As shown in the subsequent sections, the range of objective functions $F_r$, $F_a$, $F_n$ (defined later) is \([0, 1]\); thus we use the weights \(\lambda_i\) for determining the relative importance of one objective function over another.

\if 0
\begin{equation*}
cos(\vec{x}, \vec{y}) = \frac{\vec{x} \cdot \vec{y}}{\left|\vec{x} \right| \left|\vec{y} \right|} = \frac{\sum_{k=1}^{d} x_{k} \times y_{k}}{\sqrt{\sum_{k=1}^{d} x_{k}^2} \sqrt{\sum_{k=1}^{d} y_{k}^2}}
\end{equation*}

\fi

\subsubsection{Repulsion}
For any word \(w  \in  V_{d}\), we aim to minimize the  gender bias based illicit associations. Therefore, our objective function aims to `repel' \(\vec{w}_{d}\) from the neighbouring word vectors which have a high value of indirect bias ($\beta$) with it. Consequently, we name it `repulsion' (\(F_{r}\)) and primarily define the repulsion set \(S_{r}\) to be used in \(F_{r}\) as follows.
\newtheorem{definition}{Definition}
\theoremstyle{definition}
\begin{definition}
For a given word $w$, the repulsion set \(S_{r}\) is defined as $
S_{r} = \{n_{i}: n_{i} \in N_{w} \ \text{and} \  \beta(\vec{w}, \vec{n_{i}})\ > \ \theta_{r} \}$,
where $N_{w}$ is the set of top 100 neighbours obtained from the original word vector $\vec{w}$.
\end{definition}
Since we aim to reduce the unwanted semantic similarity between \(\vec{w}_d\) and the set of vectors \(S_r\), we define the objective function \(F_{r}\) as follows.
\begin{equation*}
\begin{aligned}
 F_r(\vec{w}_d) &= \left( \sum_{n_i \epsilon S_r}  \bigg | cos( \vec{w}_d, \vec{n_i} ) \bigg | \right) \bigg/ \left| S_r \right |, \\
 F_r(\vec{w}_d) & \in [0,1] 
 \end{aligned}
\end{equation*}
 For our experiments, we find that $\theta_r=0.05$ is the appropriate threshold to repel majority of gender-biased neighbours.

\subsubsection{Attraction}
For any word $w \in V_{d}$,  we aim to minimize the loss of semantic and analogical properties for its debiased counterpart $\vec{w}_{d}$. Therefore, our objective function aims to attract $\vec{w_{d}}$ towards $\vec{w}$ in the word embedding space. Consequently, we name it `attraction' ($F_{a}$) and define it as follows:
\begin{align*}
    F_{a}(\vec{w}_d) &= |\cos(\vec{w}_{d},\vec{w}) -\cos(\vec{w},\vec{w})|/2\\
    &= |\cos(\vec{w}_{d},\vec{w}) - 1|/2
, F_{a}(\vec{w}_d) \in [0,1]
\end{align*}

\subsubsection{Neutralization} 
For any word \(w \in V_{d}\), we aim to minimize its bias towards any particular gender. Therefore, the objective function \(F_n\) represents the absolute value of dot product of word vector $\vec{w}_{d}$ with the gender direction $\vec{g}$ (as defined by \citet{bolukbasi2016man}). Consequently, we name it `neutralization' ($F_n$) and define it as follows:
\begin{equation*}
  F_{n}(\vec{w}_d) =  |cos(\vec{w}_{d},\vec{g})|, F_{n} \in [0, 1]
\end{equation*}

\begin{figure*}[t]
\centering

\begin{subfigure}[b]{0.85\textwidth}
\includegraphics[width = \linewidth]{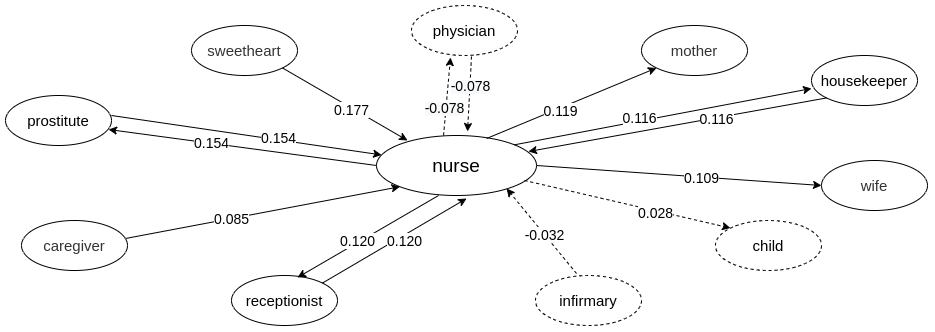}
\caption{A sub-graph of BBN with respect to the word `nurse'}
\label{fig:1}
\end{subfigure}

\begin{subfigure}[b]{0.18\textwidth}
\includegraphics[width=\textwidth]{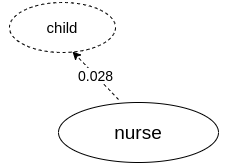}
\caption{Edge with $\beta$ < $\theta_{s}$}
\label{fig:2}
\end{subfigure}
\hfill
\begin{subfigure}[b]{0.21\textwidth}
\includegraphics[width=\textwidth]{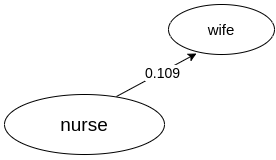}
\caption{Outgoing edge}
\label{fig:3}
\end{subfigure}
\hfill
\begin{subfigure}[b]{0.18\textwidth}
\includegraphics[width=\textwidth]{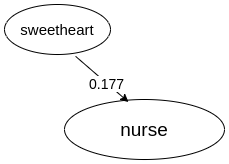}
\caption{Incoming edge}
\label{fig:4}
\end{subfigure}
\hfill
\begin{subfigure}[b]{0.20\textwidth}
\includegraphics[width=\textwidth]{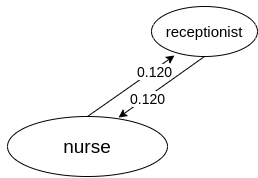}
\caption{Dual edges}
\label{fig:5}
\end{subfigure}

\caption{(\subref{fig:1}): A sub-graph of BBN formed by Algorithm \ref{algo:bbn} for GloVe  \cite{pennington2014GloVe} trained on  2017-January  dump  of  Wikipedia; we discuss the structure of the graph with respect to the word `nurse'. We illustrate four possible scenarios with respect to their effect on GIPE, with $\theta_{s}$ = 0.05: (\subref{fig:2})  An edge with $\beta$ < $\theta_{s}$ may not contribute to $\gamma_{i}$ or $\eta_{w_{i}}$; (\subref{fig:3})  An outgoing edge may contribute to $\eta_{w_{i}}$ only; (\subref{fig:4}) An incoming edge may contribute to $\gamma_{i}$ only; (\subref{fig:5}) Incoming and outgoing edges may contribute to $\gamma_{i}$ and $\eta_{w_{i}}$ respectively. Every node pair association can be categorized as one of the aforementioned four cases.}
\label{fig : BBN}
\end{figure*}

\subsubsection{Time complexity of RAN-Debias}
Computationally, there are two major components of RAN-Debias:
\begin{enumerate}[nolistsep]
    \item Calculate neighbours for each word $w \in V_d$ and store them in a hash table. This has a time complexity of $O(n^2)$ where $n=|V_d|$.
    \item Debias each word using gradient descent, whose time complexity  is $O(n)$.
\end{enumerate}
The overall complexity of RAN-Debias is $O(n^2)$ i.e., quadratic with respect to the cardinality of debias set $V_d$.

\subsection{Gender-based Illicit Proximity Estimate (GIPE)}

In Section \ref{typesGB}, we defined the gender proximity bias ($\eta$). In this section, we extend it to the embedding level for generating a global estimate. Intuitively, an estimate can be generated by simply taking the mean of $\eta_{w}$, $\, \forall w \in V_{d}$. However, this computation assigns equal importance to all $\eta_{w}$ values, which is an oversimplification. A word $w$ may itself be in the proximity of another word $w^{'} \in V_{d}$ through gender-biased associations, thereby increasing $\eta_{w^{'}}$. Such cases in which $w$ increases $\eta_{w^{'}}$ for other words should also be taken into account. Therefore, we use a weighted average of $\eta_{w}$, $\forall w  \in V$ for determining a global estimate. \textcolor{blue}{We first define a weighted directed network, called \textbf{Bias Based Network (BBN)}. The use of a graph data structure makes it easier to understand the intuition behind GIPE.}

\theoremstyle{definition}
\begin{definition}
Given a set of non gender-specific words $W$, bias based network is a directed graph $G = (V,E)$, \textcolor{blue}{where nodes represent word vectors and weights of directed edges represent the indirect bias $\left(\beta\right)$ between them.} The vertex set $V$ and edge set $E$ are obtained according to Algorithm \ref{algo:bbn}.
\end{definition}

\begin{algorithm}[h]\small
\DontPrintSemicolon
\SetKwInOut{Input}{Input}
\SetKwInOut{Output}{Output}
\Input{$\xi$: word embedding set, \newline
                 $W$: set of non gender-specific words,\\
                 $n$: number of neighbours}
\Output{$G$: bias based network}
$V = [ \ ]$, $E = [ \ ]$\\
 \For{$x_{i}\ \in\ W$}{
 $N \ = \  \underset{\xi' : |\xi'| = n}{\text{argmax}} \left(\cos(\vec{x_{i}}, \vec{x_{k}}): x_{k} \in \xi' \subseteq \xi\right)$ \newline
 $V.insert(x_{i})$\\
\For{$x_{k}\ \in\ N$}{
 $E.insert\left( x_{i},x_{k},\beta \left(\vec{x_{i}},\vec{x_{k}}\right) \right)$\\
 $V.insert\left(x_{k}\right)$
}
}
G = (V, E) \\
\Return{$G$}
\caption{Compute BBN for the given set of word vectors}\label{algo:bbn}
\label{algo:max}
\end{algorithm}

For each word $w_i$ in $W$, we find $N$, the set of top $n$  word vectors having the highest cosine similarity with $\vec{w_i}$ (we keep $n$ to be 100 to reduce computational overhead without comprising on quality). For each pair $(\vec{w_i}, \vec{w_k})$, where $w_k \in N$, a directed edge is assigned from $w_i$ to $w_k$ with the edge weight being $\beta(\vec{w_i}, \vec{w_k})$. In case the given embedding is a debiased version, we use the non-debiased version of the embedding for computing $\beta(\vec{w_i},\vec{w_k})$. Figure \ref{fig:1} portrays a sub-graph in BBN. By representing the set of non gender-specific words as a weighted directed graph we can use the number of outgoing and incoming edges for a node (word $w_{i}$) for determining $\eta_{w_{i}}$ and its weight respectively, thereby leading to the formalization of GIPE as follows.
\theoremstyle{definition}
\begin{definition}\label{def:GIPEdef}
For a BBN $G$, the Gender-based Illicit Proximity Estimate of G, indicated by $GIPE(G)$ is defined as:
\begin{equation*}
    GIPE(G) = \frac{\sum _{i=1}^{|V|}\gamma_{i}\eta_{w_{i}}}{\sum _{i=1}^{|V|}\gamma_{i}}
\end{equation*}
where, for a word $w_{i}$, $\eta_{w_{i}}$ is the gender-based proximity bias as defined earlier, $\epsilon$\;\ is a (small) positive constant, and  $\gamma_{i}$  is the weight, defined as:
\begin{equation}\label{gamma}
\resizebox{0.9\hsize}{!}{
$\gamma_{i} = 1 + \frac{|\{v_i: ( v_{i},w_{i} ) \ \in \ E, \beta \left(\vec{v_{i}},\vec{w_{i}}\right) > \ \theta_s\}|}{\epsilon+ |\{v_i: ( v_{i},w_{i} ) \ \in \ E\}|}$
}%
\end{equation}
\end{definition}

The intuition behind the metric is as follows. For a bias based network $G$, $GIPE(G)$ is the weighted average of gender-based proximity bias (\(\eta_{w_{i}}\)) for all nodes $w_{i} \ \in \ W$, where the weight of a node is $\gamma_{i}$, which signifies the importance of the node in contributing towards the gender-based proximity bias of other word vectors. $\gamma_{i}$ takes into account the number of incoming edges having $\beta$ higher than a threshold \(\theta_s\). Therefore, we take into account how the neighbourhood of a node contributes towards illicit proximities (having high $\beta$ values for outgoing edges) as well as how a node itself contributes towards illicit proximities of other nodes (having high $\beta$ values for incoming edges). For illustration, we analyze a sub-graph in Figure \ref{fig : BBN}. By incorporating $\gamma_{i}$, we take into account both dual (Figure \ref{fig:5}) and incoming  (Figure \ref{fig:4}) edges, which would not have been the case otherwise. In GloVe (2017-January dump of Wikipedia), the word `sweetheart' has `nurse' in the set of its top 100 neighbours and $\beta$ > $\theta_{s}$; however, `nurse' does not have `sweetheart' in the set of its top 100 neighbours. Hence, while `nurse' contributes towards gender-based proximity bias of the word `sweetheart', vice versa is not true. Similarly, if dual-edge exists, then both $\gamma_{i}$ and $\eta_{w_{i}}$ are taken into account. Therefore, GIPE considers all possible cases of edges in BBN, making it a holistic metric.

\section{Experiment Results}
We conduct the following  performance evaluation tests:

\noindent$\bullet$ We compare KBC with SVM based \cite{bolukbasi2016man} and RIPA based \cite{ethayarajh2019understanding} methods for word classification.
\\
\noindent$\bullet$ We evaluate the capacity of RAN-Debias on GloVe ({\em aka} RAN-GloVe) for the gender relational analogy dataset -- SemBias \cite{zhao2018learning}.
\\
\noindent$\bullet$ We demonstrate the ability of  RAN-GloVe to mitigate gender proximity bias by computing and contrasting the GIPE value.
\\
\noindent$\bullet$  We evaluate RAN-GloVe on several benchmark datasets for similarity and analogy tasks, showing that RAN-GloVe introduces minimal semantic offset to ensure  quality of the word embeddings.
\\
\noindent$\bullet$ We demonstrate that RAN-GloVe successfully mitigates gender bias in a downstream application - coreference resolution.

\textcolor{blue}{Although we report and analyze the performance of RAN-GloVe in our experiments, we also applied RAN-Debias to other popular non-contextual and monolingual word embedding, Word2vec  \cite{mikolov2013efficient} to create RAN-Word2vec. As expected, we observed similar results (hence not reported for the sake of brevity), emphasizing the generality of RAN-Debias.} Note that the percentages mentioned in the rest of the section are relative unless stated otherwise.

\subsection{Training Data and Weights}\label{sec:exp-weights}
We use GloVe \cite{pennington2014GloVe} trained on 2017-January dump of Wikipedia, consisting of $322,636$ unique word vectors of $300$ dimensions. We apply KBC on the vocabulary set $V$ obtaining $V_p$ and $V_d$ of size 47,912 and 274,724 respectively. Further, judging upon the basis of performance evaluation tests as discussed above, we experimentally select the weights in Equation \ref{eq:objective} as \(\lambda_1= 1/8,\lambda_2=6/8, \text{ and } \lambda_3=1/8\). 

\subsection{Baselines for Comparisons}

We compare RAN-GloVe against the following word embedding models, each of which is trained on the 2017-January dump of Wikipedia.
\\
\noindent$\bullet$ \textbf{GloVe}: A pre-trained word embedding model as mentioned earlier. This baseline represents the non-debiased version of word embeddings.
\\
\noindent$\bullet$  \textbf{Hard-GloVe}: Hard-Debias GloVe; we use the debiasing method\footnote{\url{https://github.com/tolga-b/debiaswe}} proposed by \citet{bolukbasi2016man} on GloVe.
\\
\noindent$\bullet$ \textbf{GN-GloVe}: Gender-neutral GloVe; we use the original\footnote{\url{https://github.com/uclanlp/gn\_GloVe}} debiased version of GloVe released by \citet{zhao2018learning}.
\\
\noindent$\bullet$ \textbf{GP-GloVe}: Gender-preserving GloVe; we use the original\footnote{\url{https://github.com/kanekomasahiro/gp\_debias}} debiased version of GloVe released by \citet{kaneko2019gender}.

\subsection{Word Classification}\label{classif}

 We compare KBC with RIPA based (unsupervised) \cite{ethayarajh2019understanding} and SVM based (supervised)  \cite{bolukbasi2016man} approaches for word classification. We create a balanced labeled test set consisting of a total of $704$ words, with $352$ words for each category -- gender-specific and non gender-specific. For the non gender-specific category,  we select all the 87 neutral and biased words from the SemBias dataset \cite{zhao2018learning}.
 Further, we select all 320, 40 and 60 gender-biased occupation words released by \citet{bolukbasi2016man, zhao2018gender} and \citet{winogender} respectively. After combining and removing duplicate words from them, we obtain 352 non gender-specific words. For the gender-specific category, we use a list of 222 male and 222 female words  provided by \citet{zhao2018learning}. We use stratified sampling to under-sample 444 words into 352 words for balancing the classes. The purpose of creating this diversely sourced dataset is to provide a robust ground-truth for evaluating the efficacy of different word classification algorithms. 

 Table \ref{table:kfbc} shows precision, recall, F1-score, AUC-ROC, and accuracy by considering gender-specific words as the positive class and non gender-specific words as the negative class. Thus, for KBC, we consider the output set $V_{p}$ as the positive and $V_{d}$ as the negative class.
  
 The SVM based approach achieves high precision but at the cost of a low recall. While the majority of the words classified as gender-specific are correct, it achieves this due to the limited coverage of the rest of gender-specific words, resulting in them being classified as non gender-specific, thereby reducing the recall drastically.
  
The RIPA approach performs fairly with respect to precision and recall.  Unlike SVM, RIPA is not biased towards a particular class and results in rather fair performance for both the classes.  
Almost similar to SVM, KBC also correctly classifies most of the gender-specific words but in an exhaustive manner, thereby leading to much fewer misclassification of gender-specific words as non gender-specific. As a result, KBC achieves sufficiently high recall.
  
 Overall, KBC outperforms the best baseline by an improvement of 2.7\% in AUC-ROC, 15.6\% in F1-score, and 9.0\% in the accuracy. \textcolor{blue}{Additionally,  since KBC entirely depends on knowledge bases, the absence of a particular word in them may result in misclassification. This could be the reason behind  the lower precision of KBC as compared to SVM-based classification and can be improved upon by incorporating more extensive knowledge bases.}

  \begin{table}[t]
    \centering
    \scalebox{0.85}{
    \begin{tabular}{|l| c c c c c |}
    \hline
    \textbf{Method} & \textbf{Prec} & \textbf{Rec} & \textbf{F1} & \textbf{AUC-ROC} & \textbf{Acc}\\
    \hline
    SVM    & \textbf{97.20} & 59.37   & 73.72   & 83.98  & 78.83 \\
    RIPA   & 60.60          & 53.40   & 56.79   & 59.51  & 59.35\\
    \hline
    KBC    & 89.65        & \textbf{81.25}   & \textbf{85.24} & \textbf{86.25} & \textbf{85.93}\\
    \hline
    \end{tabular} }
    \caption{Comparison between our proposed method (KBC), RIPA \cite{ethayarajh2019understanding} and SVM \cite{bolukbasi2016man} based word classification methods via precision (Prec), recall (Rec), F1-score (F1), AUC-ROC and accuracy (Acc). }
    \label{table:kfbc}
\end{table}

\begin{table}[t]\label{sembias}
    \centering
    \scalebox{0.72}{
    \begin{tabular}{|l | l |c c c |}
        \hline
        
          \textbf{Dataset} & \textbf{Embedding} & \textbf{Definition} $\uparrow$  & \textbf{Stereotype} $\downarrow$  & \textbf{None} $\downarrow$ \\
        \hline
          \multirow{5}{1.5 cm}{SemBias}
          
          & GloVe & 80.2& 10.9& 8.9\\
          & Hard-GloVe & 84.1 & 6.4 & 9.5 \\
          & GN-GloVe& \bf 97.7& 1.4& \bf 0.9\\
          & GP-GloVe & 84.3& 8.0& 7.7\\
          
          \cline{2-5}
          & RAN-GloVe & 92.8 & \textbf{1.1} & 6.1 \\
           \hline
        \end{tabular}}
        \caption{Comparison for the gender relational analogy test on the SemBias dataset. $\uparrow$ ($\downarrow$) indicates that higher (lower) value is better.}
    \label{tab:embeds}
\end{table}

\begin{table}[!t]
    \centering
    \scalebox{0.8}{
    \begin{tabular}{|>{\centering\arraybackslash}p{0.8cm} | l|c c c|}
    \hline
         \multirow{2}{3em}{\textbf{Input}} & \multirow{2}{3em}{\textbf{Embedding}} & 
         \multicolumn{3}{c|}{\textbf{GIPE}}
         \\
         & & $\theta_s=0.03$ & $\theta_s = 0.05$ & $\theta_s=0.07$  \\
       \hline
       \multirow{5}{1.0 cm}{$V_{d}$}
          & GloVe & 0.115 & 0.038 & 0.015 \\
        & Hard-GloVe  & 0.069 & 0.015 & 0.004 \\
          & GN-GloVe & 0.142 & 0.052 & 0.022\\
          & GP-GloVe & 0.145 & 0.048 & 0.018\\
         \cline{2-5}
         & RAN-GloVe & \textbf{0.040} & \textbf{0.006} & \textbf{0.002}\\
       \hline
       \cline{1-5}
       \multirow{5}{1.0 cm}{$H_{d}$} 
         & GloVe & 0.129 & 0.051 & 0.024 \\
        & Hard-GloVe  & 0.075 & 0.020 & \textbf{0.007} \\
         & GN-GloVe & 0.155 & 0.065 & 0.031\\
         & GP-GloVe & 0.157 & 0.061 & 0.027\\
         \cline{2-5}
         & RAN-GloVe & \textbf{0.056} & \textbf{0.018} & 0.011\\
    \hline
    \end{tabular}
    }
    \caption{GIPE (range: 0-1) for different values of $\theta_s$ (lower value is better).}
    \label{tab:GIPE}
\end{table}

\subsection{Gender Relational Analogy}
\label{sec:sembias}

To evaluate the extent of gender bias in RAN-GloVe, we perform gender relational  analogy test on the SemBias \cite{zhao2018learning} dataset.  Each instance of SemBias contains four types of word pairs: a gender-definition word pair (\textbf{Definition}; `headmaster-headmistress'), a gender-stereotype word pair (\textbf{Stereotype}; `manager-secretary') and two other word pairs which have similar meanings but no gender-based relation (\textbf{None}; `treble - bass'). There are a total of 440 instances in the semBias dataset, created by the cartesian product of 20 gender-stereotype word pairs and 22 gender-definition word pairs. From each instance, we select a word pair $(a,b)$ from the four word pairs such that using the word embeddings under evaluation, cosine similarity of the word vectors ($\vec{he}-\vec{she}$) and ($\vec{a}-\vec{b}$) would be maximum. Table \ref{tab:embeds} shows an embedding-wise comparison on the SemBias dataset. The accuracy is measured in terms of the percentage of times each type of word pair is selected as the top for various instances. RAN-GloVe outperforms all other post-processing debiasing methods by achieving at least 9.96\% and 82.8\% better accuracy in gender-definition and gender-stereotype, respectively. \textcolor{blue}{We attribute this performance to be an effect of superior vocabulary selection by KBC and the neutralization objective of RAN-Debias. KBC classifies the words to be debiased or preserved with high accuracy, while the neutralization objective function of RAN-Debias directly minimizes the preference of a biased word between `he' and `she'; reducing the gender cues that give rise to unwanted gender-biased analogies (Table \ref{tab:ablation-sembias}). Therefore, although RAN-GloVe achieves lower accuracy for gender-definition type as compared to (learning-based) GN-GloVe, it outperforms the next best baseline in \textbf{Stereotype} by at least 21.4\%}.

\subsection{Gender-based Illicit Proximity Estimate}

GIPE analyses the extent of undue gender bias based proximity between word vectors. An embedding-wise comparison for various values of $\theta_s$ is presented in Table \ref{tab:GIPE}. For a fair comparison, we compute GIPE for a BBN created upon our debias set $V_d$ as well as for $H_d$, the set of words debiased by \citet{bolukbasi2016man}. 

Here, $\theta_{s}$ represents the threshold as defined earlier in Equation \ref{gamma}. As it may be inferred from Equations \ref{eqn:gbpb} and \ref{gamma}, upon increasing the value of $\theta_{s}$, for a word $w_{i}$, the value of both $\eta_{w_{i}}$ and $\gamma_{i}$ decreases, as lesser number of words qualify the threshold for selection in each case. Therefore, as evident from Table \ref{tab:GIPE}, value of GIPE decreases with the increase of $\theta_{s}$.

\begin{table*}[t]
    \centering
    \scalebox{0.9}{
    \begin{tabular}{|l|c c c || c c c c c c|}
        \hline
         \multirow{2}{6em}{\textbf{Embedding}} & \multicolumn{3}{c||}{\textbf{(a) Analogy}} & \multicolumn{6}{c|}{\textbf{(b) Semantic}}  \\ \cline{2-10}
          & Google-Sem & Google-Syn & MSR  & RG & MTurk & RW & MEN & SimLex999 & AP \\
        \hline
          GloVe & 79.02 & 52.26 & 51.49 & 75.29 & 64.27 & 31.63 & 72.19 & 34.86 & 60.70 \\
          Hard-GloVe & 80.26 & 62.76 & \textbf{51.59} & \textbf{76.50} & 64.26 & 31.45 & 72.19 & 35.17 & 59.95 \\ 
          GN-GloVe  & 76.13 & 51.00 & 49.29 &74.11 &\textbf{66.36} & \textbf{36.20} &\textbf{74.49} &\textbf{37.12}& 61.19\\
          GP-GloVe  & 79.15 & 51.55 & 48.88  & 75.30 & 63.46 & 27.64 & 69.78 & 34.02 &57.71\\
          \hline
          RAN-GloVe  & \textbf{80.29} & \textbf{62.89} & 50.98  & 76.22 & 64.09 & 31.33 & 72.09 & 34.36 & \textbf{61.69}\\
        \hline
    \end{tabular}}
    
    \caption{Comparison of various embedding methods for (a) analogy tests (performance is measured in accuracy) and (b) word semantic similarity tests (performance is measured in terms of  Spearman rank correlation).}
    \label{tab:semanal}
\end{table*}

For the input set $V_{d}$, RAN-GloVe outperforms the next best baseline (Hard-GloVe) by at least $42.02\%$. We attribute this to the inclusion of the repulsion objective function $F_{r}$ in Equation \ref{eqn:obj}, which reduces the unwanted gender-biased associations between words and their neighbours. For the input set $H_{d}$, RAN-GloVe performs better than other baselines for all values of $\theta_{s}$ except for $\theta_{s} = 0.07$ where it closely follows Hard-GloVe.

Additionally, $H_{d}$ consists of many misclassified gender-specific words, as observed from the low recall performance at the word classification test in Section \ref{classif}. Therefore, the values of GIPE corresponding to every value of $\theta_{s}$ for the input $H_{d}$ is higher as compared to the values for $V_{d}$. 

\textcolor{blue}{Although there is a significant reduction in GIPE value for RAN-GloVe as compared to other word embedding models, word pairs with noticeable $\beta$ values still exist (as indicated by non-zero GIPE values), which is due to the trade-off between semantic offset and bias reduction. As a result, GIPE for RAN-GloVe is not perfectly zero but close to it.}

\subsection{Analogy Test}
The task of analogy test is to answer the following question: ``p is to q as r is to ?". Mathematically, it aims at finding a word vector $\vec{w}\textsubscript{s}$ which has the maximum cosine similarity with ($\vec{w}\textsubscript{q} - \vec{w}\textsubscript{p} + \vec{w}\textsubscript{r}$). \textcolor{blue}{However, \citet{schluter-2018-word} highlights some critical issues with word analogy tests. For instance, there is a mismatch between the distributional hypothesis used for generating word vectors and the word analogy hypothesis. Nevertheless, following the practice of using word analogy test  to ascertain the semantic prowess of word vectors, we evaluate RAN-GloVe to provide a fair comparison with other baselines.}

We use  Google \cite{mikolov2013efficient} (semantic  (Sem) and syntactic (Syn) analogies, containing a total 19,556 questions) and MSR \cite{mikolov2013linguistic} (containing a total 7,999 syntactic questions) datasets for evaluating the performance of word embeddings. We use  3\textsc{Cos}\textsc{Mul}\cite{levy2014linguistic} for finding $\vec{w}\textsubscript{s}$.

Table \ref{tab:semanal}(a) shows that RAN-GloVe  outperforms other baselines on the Google (Sem and Syn) dataset while closely following on the MSR dataset. The improvement in performance can be attributed to the removal of unwanted neighbours of a word vector (having gender bias based proximity), while enriching the neighbourhood with those having empirical utility, leading to a better performance in analogy tests.

\subsection{Word Semantic Similarity Test}
Word semantic similarity task is a measure of how closely a word embedding model captures the similarity between two words as compared to human-annotated ratings.
For a word pair, we compute the cosine similarity between the word embeddings and its Spearman correlation with the human ratings. The word pairs are selected from the following benchmark datasets:
RG \cite{rubenstein1965contextual},
MTurk \cite{radinsky2011word}, RW \cite{luong2013better},
MEN \cite{bruni2014multimodal}, SimLex999 \cite{hill2015simlex} and  AP \cite{almuhareb2005concept}. The results for these tests are obtained from the word embedding benchmark package \cite{jastrzebski2017evaluate}\protect\footnote{\url{https://github.com/kudkudak/word-embeddings-benchmarks}}. Note that it is not our primary aim to achieve a state-of-the-art result in this test. It is only considered to evaluate  semantic loss. Table \ref{tab:semanal}(b) shows that RAN-GloVe  performs better or follows closely to the best baseline. This shows that RAN-Debias introduces minimal semantic disturbance.

\subsection{Coreference Resolution}
Finally, we evaluate the performance of RAN-GloVe on a downstream application task -- coreference resolution.
The aim of coreference resolution is to identify all expressions which refer to the same entity in a given text. We evaluate the embedding models on the OntoNotes 5.0 \cite{weischedel2012ontonotes} and the WinoBias \cite{zhao2018gender}  benchmark datasets. 
% WinoBias reveals the extent of gender bias present in coreference resolution systems. It 
\textcolor{blue}{WinoBias comprises of sentences constrained by two prototypical templates (Type 1 and Type 2), where each template is further divided into two subsets (PRO and ANTI). Such a construction facilitates in revealing the extent of gender bias present in coreference resolution models. While both templates are designed to assess the efficacy of coreference resolution models, Type 1 is exceedingly challenging as compared to Type 2 as it has no syntactic cues for disambiguation. Each template consists of two subsets for evaluation -- pro-stereotype (PRO) and anti-stereotype (ANTI). PRO consists of sentences in which the gendered pronouns refer to occupations biased towards the same gender. For instance, consider the sentence ``The doctor called the nurse because he wanted a vaccine.'' Stereotypically, `doctor' is considered to be a male-dominated profession, and the gender of pronoun referencing it (`he') is also male. Therefore, sentences in PRO are consistent with societal stereotypes. ANTI consists of the same sentences as PRO, but the gender of the pronoun is changed. Considering the same example but by replacing `he' with `she', we get: ``The doctor called the nurse because she wanted a vaccine.'' In this case, the gender of pronoun (`she') which refers to `doctor' is female. Therefore, sentences in ANTI are not consistent with societal stereotypes.
Due to such construction, gender bias in the word embeddings used for training the coreference model would naturally perform better in PRO than ANTI and lead to a higher absolute difference ({\em Diff}) between them. While a lesser gender bias in the model would attain a smaller {\em Diff}, the ideal case produces an absolute difference of zero.}

Following the coreference resolution testing methodology used by \citet{zhao2018learning}, we train the coreference resolution model proposed by \citet{lee2017end} on the OntoNotes train dataset for different embeddings. Table \ref{tab:coref} shows F1-score on OntoNotes 5.0 test set, WinoBias PRO and ANTI test set for Type 1 template, along with the absolute difference ({\em Diff}) of F1-scores on PRO and ANTI datasets for different word embeddings. The results for GloVe, Hard-GloVe, and GN-GloVe are obtained from \citet{zhao2018learning}.

\begin{table}[t]
    \centering
    \scalebox{0.85}{
    \begin{tabular}{|l|c| c c |c|}
    \hline
         \textbf{Embedding} & \textbf{OntoNotes} & \textbf{PRO} & \textbf{ANTI} & \textbf{\textit{Diff}} \\
       \hline

         GloVe     &\textbf{66.5} &\textbf{76.2} & 46.0 & 30.2 \\
         Hard-GloVe &66.2 & 70.6 & 54.9 & 15.7\\
         GN-GloVe  &66.2 & 72.4 & 51.9 & 20.5 \\
         GP-GloVe  &66.2 &70.9 &52.1 &18.8\\
         \hline
         RAN-GloVe  &66.2 &61.4 &\textbf{61.8} &\textbf{0.4}\\
    \hline
    \end{tabular} }
    \caption{F1-Score (in \%) in the task of coreference resolution. {\em Diff} denotes the absolute difference between F1-score on PRO and ANTI datasets.}
    \label{tab:coref}
\end{table}

Table \ref{tab:coref} shows that RAN-GloVe achieves the smallest absolute difference between scores on PRO and ANTI subsets of WinoBias, significantly outperforming other embedding models and achieving 97.4\% better {\em Diff} (see Table \ref{tab:coref} for the definition of {\em Diff}) than the next best baseline (Hard-GloVe) and 98.7\% better than the original GloVe. \textcolor{blue}{This lower {\em Diff} is achieved by an improved accuracy in ANTI and a reduced accuracy in PRO. We hypothesise that the high performance of non-debiased GloVe in PRO is due  to the unwanted gender cues rather than the desired coreference resolving ability of the model. Further, the performance reduction in PRO for the other debiased versions of GloVe also corroborates this hypothesis. Despite debiasing GloVe, a considerable amount of gender cues  remain in the baseline models as quantified by a lower, yet significant {\em Diff}. In contrast, RAN-GloVe is able to remove gender cues dramatically, thereby achieving an almost ideal {\em Diff}.}  Additionally, the performance of RAN-GloVe on the OntoNotes 5.0 test set is comparable with that of other embeddings.

\subsection{Ablation Study}
To quantitatively and qualitatively analyze the effect of neutralization and repulsion in RAN-Debias, we perform an ablation study. We examine the following changes in RAN-Debias independently:
\begin{enumerate}[nolistsep]
    \item Nullify the effect of repulsion by setting $\lambda_1=0$, thus creating AN-GloVe. 
    \item  Nullify the effect of neutralization by setting $\lambda_3=0$, thus creating RA-GloVe.
\end{enumerate}
We demonstrate the effect of the absence of neutralization or repulsion through a comparative analysis on GIPE and the SemBias analogy test.

\begin{table}[h]
    \centering
    \scalebox{0.8}{
    \begin{tabular}{|>{\centering\arraybackslash}p{0.8cm} | l|c c c|}
    \hline
         \multirow{2}{3em}{\textbf{Input}} & \multirow{2}{3em}{\textbf{Embedding}} & 
         \multicolumn{3}{c|}{\textbf{GIPE}}
         \\
         & & $\theta_s=0.03$ & $\theta_s = 0.05$ & $\theta_s=0.07$  \\
       \hline
       \multirow{3}{1.0 cm}{$V_{d}$}
          & AN-GloVe & 0.069 & 0.015 & 0.004 \\
        & RA-GloVe  & 0.060 & 0.014 & 0.007 \\
        & RAN-GloVe  & \textbf{0.040} & \textbf{0.006} & \textbf{0.002} \\\hline
    \end{tabular}}
    \caption{Ablation study - GIPE for AN-GloVe and RA-GloVe.}
    \label{tab:ablation-GIPE}
\end{table}

The GIPE values for AN-GloVe, RA-GloVe, and RAN-GloVe are presented in Table \ref{tab:ablation-GIPE}. We observe that in the  absence of repulsion (AN-GloVe), the performance is degraded by at least 72\% compared to RAN-GloVe. It indicates the efficacy of repulsion in our objective function as a way to reduce the unwanted gender-biased associations between words and their neighbours, thereby reducing  GIPE. Further, even in the absence of neutralization (RA-GloVe), GIPE is worse by at least 50\% as compared to RAN-GloVe. In fact, the minimum GIPE is observed for RAN-GloVe, where both repulsion and neutralization are used in synergy as compared to the absence of any one of them.

\begin{table}[t]
    \centering
    \scalebox{0.75}{
    \begin{tabular}{|>{\centering\arraybackslash}p{1.1cm} | p{1.6cm}|c c c|}
    \hline
         \multirow{2}{3em}{\textbf{Word}} & \multirow{2}{3em}{\textbf{Neighbour}} & 
         \multicolumn{3}{c|}{\textbf{Embedding}}
         \\
         & & AN-GloVe & RA-GloVe & RAN-GloVe  \\
       \hline
        \multirow{2}{1.0 cm}{Captain}
          & sir & 28 & 22 & 52 \\
        & james  & 26 & 30 & 75 \\
       \hline
       \multirow{2}{1.0 cm}{Nurse}
          & women & 57 & 56 & 97 \\
        & mother  & 49 & 74 & 144 \\
       \hline
       \multirow{2}{1.0 cm}{Farmer}
          & father & 22 & 54 & 86 \\
        & son  & 45 & 90 & 162 \\
    \hline
    \end{tabular}}
    \caption{\textcolor{blue}{For three professions, we compare the ranks of their neighbours  due to illicit proximities (the values denote the ranks).}}
    \label{tab:ablation-rank}
\end{table}

\textcolor{blue}{To illustrate further,  Table \ref{tab:ablation-rank} shows the rank of neighbours having illicit proximities for three professions, using different version of debiased embeddings. It can be observed that the ranks in RA-GloVe are either close to or further away from the ranks in AN-GloVe, highlighting the importance of repulsion in the objective function.  Further, the ranks in RAN-GloVe are the farthest, corroborating the minimum value of GIPE as observed in Table \ref{tab:ablation-GIPE}.}

\begin{table}[h]
    \centering
    \scalebox{0.72}{
    \begin{tabular}{|l | l |c c c |}
        \hline
        
          \textbf{Dataset} & \textbf{Embedding} & \textbf{Definition} $\uparrow$  & \textbf{Stereotype} $\downarrow$  & \textbf{None} $\downarrow$ \\
        \hline
          \multirow{3}{1.5 cm}{SemBias}
          
          & AN-GloVe & \textbf{93.0}& \textbf{0.2}& 6.8\\
          & RA-GloVe & 83.2 & 7.3 & 9.5 \\
          & RAN-GloVe & 92.8 & 1.1  & \textbf{6.1} \\

          \hline
        \end{tabular}}
        \caption{Comparison for the gender relational analogy test on the SemBias dataset. $\uparrow$ ($\downarrow$) indicates that higher (lower) value is better.}
    \label{tab:ablation-sembias}
\end{table}

\begin{table*}[h]
    \centering
    \scalebox{0.9}{
    \begin{tabular}{|l | c | l | c c c c c |}
    \hline
        \multirow{2}{4em}{\textbf{Word}} & \multirow{2}{4em}{\textbf{Class}} & \multirow{2}{4em}{\textbf{Neighbour}} & \multicolumn{5}{c|}{\textbf{Embedding}}
         \\
         & & & GloVe & Hard-GloVe & GN-GloVe & GP-GloVe & RAN-GloVe
         \\
         \hline
         
         \multirow{8}{4em}{Captain} &
         \multirow{4}{4em}{A} & 
         sir & 19 & 32 & 34 & 20 & 52
         \\
         & & james & 20 & 22 & 26 & 18 & 75 
         \\
         & &  brother & 34 & 83 & 98 & 39 & 323
         \\
         & & father & 39 & 52 & 117 & 40 & 326
         \\
         
         \cline{2-8}
         & \multirow{4}{4em}{B} & 
         lieutenant & 1 & 1 & 1 & 1 & 1
         \\
         & & colonel & 2 & 2 & 2 & 2 & 2
         \\
         & & commander & 3 & 3 & 4 & 3 & 3
         \\
         & & officer & 4 & 5 & 10 & 4 & 15
         \\
         \hline
         \hline

         \multirow{8}{4em}{Nurse} &
         \multirow{4}{4em}{A} & 
         
         woman & 25 & 144 & 237 & 16 & 97
         \\
         & & mother & 27 & 71 & 127 & 25 & 144 
         \\
         & &  housekeeper & 29 & 54 & 28 & 29 & 152
         \\
         & & girlfriend & 32 & 74 & 60 & 31 & 178
         \\
         
         \cline{2-8}
         
         & \multirow{4}{4em}{B} & 
         
         nurses & 1 & 1 & 1 & 1 & 1
         \\
         & & midwife & 2 & 3 & 2 & 3 & 2
         \\
         & & nursing & 3 & 2 & 3 & 2 & 9
         \\
         & & practitioner & 4 & 5 & 4 & 5 & 3
         \\
         \hline
         \hline
         
         \multirow{8}{4em}{Socialite} & 
         \multirow{4}{4em}{A} & 
         businesswoman & 1 & 1 & 1 & 1 & 6
         \\
         & & heiress & 2 & 2 & 2 & 2 & 9 
         \\
         & &  niece & 12 & 18 & 14 & 17 & 78
         \\
         & & actress & 19 & 16 & 38 & 14 & 120
         \\
         
         \cline{2-8}
         & \multirow{4}{4em}{B} & 
         philanthropist & 3 & 3 & 3 & 3 & 1
         \\
         & & aristocrat & 4 & 4 & 4 & 4 & 3
         \\
         & & wealthy & 5 & 5 & 7 & 5 & 4
         \\
         & & socialites & 6 & 15 & 5 & 9 & 10
         \\
         \hline
         \hline
         
         \multirow{8}{4em}{Farmer} & \multirow{4}{4em}{A} & 
         father & 12 & 28 & 37 & 13 & 84
         \\
         & & son & 21 & 84 & 77 & 26 & 162 
         \\
         & &  boy & 50 & 67 & 115 & 45 & 105
         \\
         & & man & 51 & 50 & 146 & 60 & 212
         \\
         
         \cline{2-8}
         & \multirow{4}{4em}{B} & 
         rancher & 1 & 2 & 1 & 2 & 3
         \\
         & & farmers & 2 & 1 & 4 & 1 & 1
         \\
         & & farm & 3 & 3 & 5 & 4 & 2
         \\
         & & landowner & 4 & 4 & 2 & 5 & 5
         \\
         \hline
    \end{tabular}}
    \caption{For four professions, we compare the ranks of their class A and class B neighbours with respect to each embedding. Here, rank represents the position in the neighbourhood of a profession, and is shown by the values under each embedding.}
    \label{tab:word_compare}
    \vspace{-3mm}
\end{table*}

Table \ref{tab:ablation-sembias} shows that in the absence of neutralization (RA-GloVe), the tendency of favouring stereotypical analogies increases by an absolute difference of 6.2\% as compared to RAN-GloVe. On the other hand, through the presence of neutralization, AN-GloVe does not favour stereotypical analogies. This suggests that reducing the projection of biased words on gender direction through neutralization is an effective measure to reduce stereotypical analogies within the embedding space. \textcolor{blue}{For example, consider the following instance of word pairs from the SemBias dataset: \{({\em widower, widow}), ({\em book, magazine}), ({\em dog, cat}), ({\em doctor, nurse})\}, where \emph{(widower, widow)} is a gender-definition word pair while \emph{(doctor, nurse)} is a gender-stereotype word pair and the remaining are of none type as explained in Section \ref{sec:sembias}. During the evaluation, RA-GloVe incorrectly selects the gender-stereotype word pair as the closest analogy with \emph{(he, she)}, while AN-GloVe and RAN-GloVe correctly select the gender-definition word pair.} Further, we observe that RAN-GloVe is able to  maintain the high performance of AN-GloVe, and the difference is less (0.2\% compared to 1.1\%) which is compensated by the superior performance of RAN-GloVe over other metrics like GIPE.

Through this ablation study, we understand the importance of repulsion and neutralization in the multi-objective optimization function of RAN-Debias. The superior performance of RAN-GloVe can be attributed to the synergistic interplay of repulsion and neutralization. Hence, in RAN-GloVe we attain the best of both worlds.

\subsection{Case study - Neighbourhood of Words}

Here we highlight the changes in the neighbourhood (collection of words sorted in the descending order of cosine similarity with the given word) of words before and after the debiasing process. To maintain readability while also demonstrating the changes in proximity, we only analyze a few selected words. However, our proposed metric GIPE quantifies this for an exhaustive vocabulary set.

We select a set of gender-neutral professions having high values of gender-based proximity bias $\eta_{w_i}$ as defined earlier. For each of these professions, in Table \ref{tab:word_compare}, we select a set of four words from their neighbourhood for two classes:
\\
\noindent$\bullet$ \textbf{Class A}: Neighbours arising due to gender-based illicit proximities.
\\
\noindent$\bullet$ \textbf{Class B}: Neighbours whose proximities are not due to any kind of bias.

For the words in class A, the debiasing procedure is expected to increase their rank, thereby decreasing the semantic similarity, while for words belonging to class B, debiasing procedure is expected to retain or improve the rank for maintaining the semantic information.

We observe that RAN-GloVe not only maintains the semantic information by keeping the rank of words in class B close to their initial value but unlike other debiased embeddings, it drastically increases the rank of words belonging to class A. However, in some cases like the word `Socialite', we observe that the ranks of words such as `businesswoman' and `heiress', despite belonging to class A, are close to their initial values. This can be attributed to the high semantic dependence of `Socialite' on these words, resulting in a bias removal and semantic information trade-off.

\section{Conclusion}
In this paper, we proposed a post-processing gender debiasing method, called RAN-Debias. Our method not only mitigates direct bias of a word but also reduces its associations with other words that arise from gender-based predilections. We also proposed a word classification method, called KBC, for identifying the set of words to be debiased. Instead of using `biased' word embeddings, KBC uses multiple knowledge bases for word classification. Moreover, we proposed Gender-based Illicit Proximity Estimate (GIPE), a metric to quantify the extent of illicit proximities in an embedding. RAN-Debias significantly outperformed other debiasing methods on a suite of evaluation metrics, along with the downstream application task of coreference resolution while introducing minimal semantic disturbance.

In the future, we would like to enhance KBC by utilizing machine learning methods to account for the words which are absent in the knowledge base. Currently, RAN-Debias is directly applicable to non-contextual word embeddings for non-gendered grammatical languages. In the wake of recent work such as \citet{zhao-contextual}, we would like to extend our work towards contextualized embedding models and other languages with grammatical gender like French and Spanish.

\bibliography{tacl2018.bib}
\bibliographystyle{acl_natbib}
\end{document}